\documentclass[12pt,a4paper]{article}
\pdfoutput=1
\usepackage{PRIMEarxiv}

\usepackage[]{inputenc} 
\usepackage[T1]{fontenc}    
\usepackage{hyperref}       
\usepackage{url}            
\usepackage{booktabs}       
\usepackage{amsfonts}       
\usepackage{nicefrac}       
\usepackage{microtype}      
\usepackage{lipsum}         
\usepackage{graphicx}
\usepackage{subcaption}
\usepackage[square,sort,comma,numbers]{natbib}
\usepackage{doi}
\usepackage{wrapfig}
\usepackage{xcolor}
\usepackage{amsmath}
\usepackage{cleveref}       
\usepackage{float}

\title{Comparing Spectral Bias and Robustness For Two-Layer Neural Networks: SGD vs Adaptive Random Fourier Features}
\newif\ifuniqueAffiliation
\uniqueAffiliationtrue

\usepackage{authblk}

\author[1]{%
	Aku Kammonen\thanks{\texttt{akujaakko.kammonen@kaust.edu.sa}}}%
\author[2]{%
	Lisi Liang\thanks{\texttt{lisi.liang@rwth-aachen.de}}}%
\author[2]{%
	Anamika Pandey\thanks{\texttt{pandey@uq.rwth-aachen.de}}}%
\author[1,2]{%
	Ra\'{u}l Tempone\thanks{\texttt{raul.tempone@kaust.edu.sa}}}%
\affil[1]{KAUST, Saudi Arabia}
\affil[2]{RWTH Aachen, Germany}

\hypersetup{
pdftitle={Comparing Spectral Bias and Robustness For Two-Layer Neural Networks: SGD vs Adaptive Random Fourier Features},
pdfsubject={machine learning, spectral bias, adversarial attack, random features, neural network, SGD, MNIST},
pdfauthor={Aku Kammonen, Lisi Liang, Anamika Pandey},
pdfkeywords={machine learning, spectral bias, adversarial attack, random features, neural network, SGD, MNIST},
}

\begin{document}
\maketitle

\begin{abstract}
We present experimental results highlighting two key differences resulting from the choice of training algorithm for two-layer neural networks. The spectral bias of neural networks is well known, while the spectral bias dependence on the choice of training algorithm is less studied. Our experiments demonstrate that an adaptive random Fourier features algorithm (ARFF) can yield a spectral bias closer to zero compared to the stochastic gradient descent optimizer (SGD). Additionally, we train two identically structured classifiers, employing SGD and ARFF, to the same accuracy levels and empirically assess their robustness against adversarial noise attacks.
\end{abstract}



\section{Introduction}
\label{Sec:Intro}

Consider the data $\{x_n, y_n\}\in\mathbb{R}^d\times\mathbb{R}, n=1,...,N$ i.i.d. sampled from an unknown probability distribution with probability density function $\tilde{\rho}(x, y)$. We assume that there exists an unknown underlying function $f:\mathbb{R}^d\to\mathbb{R}$ such that $f(x_n) = y_n + \epsilon_n, n = 1,...,N$ where $\mathbb{E}[\epsilon_n] = 0$ and $\mathbb{E}[\epsilon_n^2] < \infty$. We address the supervised learning problem of fitting a two-layer neural network  $\beta(\boldsymbol{x}) = \sum^K_{k=1} \hat{\beta}_k e^{\mathrm{i}{\boldsymbol{\omega}_k}\cdot \boldsymbol{x}}$ to the data $\{x_n, y_n\}_{n=1}^N$ by numerically approximating a solution to the minimization problem
    \begin{equation}
    \label{eq:main_lsq_problem}
        \min_{
            (\boldsymbol{\omega}, \boldsymbol{\hat{\beta}})\in\mathbb{R}^{Kd}\times\mathbb{C}^K}\left\{\mathbb{E}_{\tilde\rho}\left[|y - \beta(x)|^2\right] + \lambda\sum_{k=1}^K|\hat{\beta}_k|^2\right\}
    \end{equation}
    where $\lambda\in\mathbb{R}$ is a Tikhonov regularization parameter. Note that, in practice, the data density $\tilde{\rho}$ is unknown, therefore we consider the empirical minimization problem instead. Considering $\boldsymbol{\omega}_k$ as i.i.d. samples from an unknown distribution with probability density function $p(\boldsymbol{\omega}_k)$ and by following \cite{Barron_1993} an $\mathcal{O}(1/K)$ error bound to Problem \eqref{eq:main_lsq_problem} is derived in \cite{kammonen2020adaptive}. 
    However, in the work \cite{kammonen2020adaptive}, only $\boldsymbol{\omega}_{k}$ is treated as a random variable, while $\hat{\beta}_{k}$ is determined for a sampled $\boldsymbol{\omega}_{k}$ without accounting for any inherent randomness in $\hat{\beta}_{k}$.
    Further, the constant in the $\mathcal{O}(1/K)$ error bound is shown to be minimized by sampling the weights $\boldsymbol{\omega}_k$ from $p_*(\boldsymbol{\omega}):=|\hat{f}(\boldsymbol{\omega})|/\|\hat{f}\|_{L^1(\mathbb{R}^d)}$ where $\hat{f}$ denotes the Fourier transform of $f$. Given only data $\{x_n, y_n\}_{n=1}^N$ means that $\hat{f}$ typically is not accessible. The goal of the adaptive random Fourier features with Metropolis sampling algorithm (ARFF), presented in and denoted as Algorithm 1 in \cite{kammonen2020adaptive}, is to approximately sample $\boldsymbol{\omega}$ from $p_*$, given only data. Applications of the ARFF algorithm include wind field reconstruction, see \cite{kiessling2021}, and pre-training of deep residual neural networks, see \cite{kammonen2022deepVSshallow}.
For simplicity in our numerical experiments, we employ the cosine activation function. This choice necessitates the introduction of a bias term $b_k$ for each $k = 1, \ldots, K$ in the argument of the cosine function. The purpose of this bias is to offset the contribution of the sinusoidal component in the Fourier representation of the target function, $f$.

For the ARFF training, the associated least squares problem, in the amplitudes $\hat{\boldsymbol{\beta}}$ for sampled frequencies $\boldsymbol{\omega}_k$ and bias $b_k, k = 1,..., K$ is $\min_{\hat{\boldsymbol{\beta}} \in \mathbb{R}^K} (N^{-1} 
    |\boldsymbol{S} \hat{\boldsymbol{\beta}} - \boldsymbol{y}|^2 + 
    \lambda |\hat{\boldsymbol{\beta}}|^2),$
where $\boldsymbol{S}\in \mathbb{R}^{N \times K}$ is a matrix with elements 
$S_{n,k} = \cos(\boldsymbol{\omega}_k \cdot \boldsymbol{x}_n + b_k)$, $n=1,...,N$, $k=1,...,K$ and 
$\boldsymbol{y} = (y_1, ..., y_N) \in \mathbb{R}^N$. 
The SGD training follows the standard approach and is adapted from the TensorFlow implementation.

Rahimi and Recht first introduce random Fourier features in \cite{rahimi2007random} as a computationally fast and accurate way to approximate the kernels of Support Vector Machines, \cite{Vladimir1998, SCSpringer2008}. In  \, \cite{tancik2020fourfeat}, they show that by manually adjusting the sampling distribution, Fourier features neural networks can learn high-frequency components and thus overcome the spectral bias. Our approach is to adaptively sample the frequencies with ARFF and quantify the spectral bias with the definition introduced in \cite{kiessling2022computable}. 

While several innovative adversarial attack methods \cite{hajri2022neural, tsai2019adversarial, andriushchenko2020square}, and
efficient defense mechanisms \cite{Friendly_noise_against_adversarial_noise, Abadi2016DeepLW, Chen2018DetectingBA} exist, we demonstrate how just changing the training algorithm can affect the robustness of a simple adversarial attack.\\
\textbf{Our contributions in summary}
\vspace{-2mm}
\begin{itemize}
    \item We show, experimentally, how training a two-layer neural network with adaptive random Fourier features \cite{kammonen2020adaptive} can overcome the spectral bias, as defined in \cite{kiessling2022computable}, compared to training with stochastic gradient descent.
    \item We show, experimentally, how two sets of identically structured neural networks with the same classification rate on the MNIST dataset have different degrees of robustness to a simple additive adversarial attack, depending on if we train them with adaptive random Fourier features \cite{kammonen2020adaptive} or with stochastic gradient descent.
\end{itemize}
\section{Spectral bias}
\label{Sec:SB}


Numerous prior investigations \cite{rahaman2019spectral, Hong2022OnTA, Wu2022ExtrapolationAS, Rahaman2018OnTS, cao2019towards}, encompassing both empirical and theoretical approaches have delved into the spectral bias of neural networks: the inclination to learn low-frequency content. Kiessling et al. introduced a computable definition of spectral bias, as documented in \cite{kiessling2022computable}, which shares similarities with the approaches outlined in \cite{xu2019frequency} and \cite{xu2019training}.

In this work, we employ the spectral bias definition provided by Kiessling et al. in \cite{kiessling2022computable} to assess and contrast the spectral bias between two-layer neural networks trained by ARFF and SGD.

The spectral bias of a neural network $\beta$ is defined as $SB = (\mathcal{E}_{high} - \mathcal{E}_{low})/(\mathcal{E}_{high} + \mathcal{E}_{low}),$ in \cite{kiessling2022computable}, where
$\mathcal{E}_{high} = \frac{\int_{\Omega_{high}} |\hat{r}_\rho(\omega)|^2d\omega}
    {(2\pi)^d Var(f(x))}$ and $\mathcal{E}_{low} = \frac{\int_{\Omega_{low}} |\hat{r}_\rho(\omega)|^2d\omega}
    {(2\pi)^d Var(f(x))}$ , respectively, represent the error in the high- and low-frequency domain. 
Here $Var(f(x))$ denotes the variance of the unknown target function $f$
and is approximated by the given dataset by using Monte Carlo integration, $\hat{r}_\rho(\omega)$ denotes the frequency spectrum of the function $r_\rho(x) = \sqrt[]{\rho(x)} (r(x)-\mathbb{E}_{\rho}[r(x)])$ where $r(x) = f(x) - \beta(x)$ is the residual, and $\rho(x)$ denotes the density of the input data $x$.

The symbols $\Omega_{\text{high}}$ and $\Omega_{\text{low}}$ denote the partitioned frequency domain, delineated by a designated cutoff frequency $\omega_{0}$, that is, $\Omega_{high} = \{\omega \in \mathbb{R}^d: |\omega|_\infty > \omega_0\}$, and $\Omega_{low} = \{\omega \in \mathbb{R}^d: |\omega|_\infty \leq \omega_0\}$. The cutoff frequency $\omega_{0}$ is specified to equalize the contributions of $\Omega_{\text{low}}$ and $\Omega_{\text{high}}$ to the total variance, expressed as
$
\label{eq:cutoffFreq}
\int_{\Omega_{\text{low}}} |\hat{f}_\rho(\omega)|^2 d\omega = \int_{\Omega_{\text{high}}} |\hat{f}_\rho(\omega)|^2 d\omega,$ where $\hat{f}_\rho$ denotes the Fourier transform of the function $f_\rho(x) = \sqrt[]{\rho(x)} (f(x)-\mathbb{E}_{\rho}[f(x)])$. The integral equality to obtain the cutoff frequencies and, consequently, the spectral bias are in practice approximated by Methods 1 and 2 given in \cite{kiessling2022computable}. 

If the neural network performs equally well for low and high-frequency content, the spectral bias ($SB$) tends to approach zero. In such cases, the neural network will be characterized as spectrally unbiased.
\vspace{-3 mm}
\subsection*{Numerical comparison of spectral bias in trained neural networks using SGD and ARFF training algorithms}
\vspace{-1 mm}
For this experiment, we consider the target function $f: \mathbb{R} \rightarrow \mathbb{R}$
\begin{gather*}
    f(x) = e^{-\frac{x^2}{2}} Si\left(\frac{x}{a}\right), \quad \text{where}~~ Si\left(x\right) = \int^{x}_0 \frac{\sin(t)}{t} dt ~~ \text{with}~~ a = 10^{-2}. 
\end{gather*}
We train a two-layer neural network $\beta(x; \boldsymbol{\omega}, \boldsymbol{\hat{\beta}}) 
    = \sum^K_{k=1} \hat{\beta}_k \cos(\omega_k\cdot x + b_k)$ to approximate the target function $f$, where $K$ is the number of nodes in the neural network, $\omega_k \in \mathbb{R}, x \in \mathbb{R}$ and $\hat{\beta} \in \mathbb{R} $. While keeping the architecture the same, we train the neural network with SGD and ARFF. We compute the spectral bias of the trained neural network after getting the desired accuracy in the approximation of the target function. We use Method 1, described in \cite{kiessling2022computable}, for this experiment. 
    
We generate training, validation, and test datasets as follows: sample $N$ $x$-points from $\mathcal{N}(0, 1)$, evaluate the target function to obtain $f(x)$, and normalize all three datasets by subtracting the mean and dividing by the standard deviation of the training dataset $\{x_{n}, f(x_{n})\}_{n=1}^N$. The error is then computed as the mean squared error for each dataset.
\begin{wrapfigure}{c}{0.45\textwidth}
	\begin{center}
		\includegraphics[width = 0.7\linewidth]{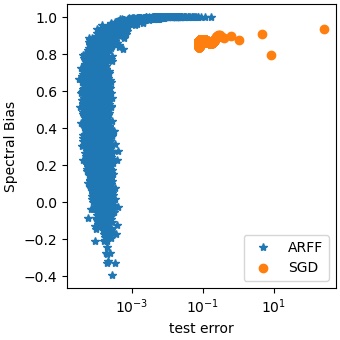}
	\end{center}
	\caption{Spectral bias comparison, computed after each epoch, of a neural network trained with SGD and ARFF by using Method 1 from \cite{kiessling2022computable}.}
	\label{Fig:SBComparison}
\end{wrapfigure}

Both training algorithms, SGD and ARFF, are executed for $M=10^4$ epochs. In the case of ARFF, we employ the following parameters: 
Tikhonov regularization $\lambda=0.05$, proposal width $\delta=2$ in the sampling algorithm, and the exponent $\gamma=3d-2$ in the acceptance-rejection ratio, where $d=1$. For SGD we utilize a batch size of $32$ and a learning rate of $0.0002$.

In Figure \ref{Fig:SBComparison}, we delineate a comparison of the spectral bias exhibited by a neural network comprising $K=1024$ neurons in the hidden layer and trained by two distinct training algorithms, SGD and ARFF.

The comparison reveals that the spectral bias of the neural network trained with SGD remains close to unity. In contrast, the ARFF training algorithm enables the neural network to achieve a state of spectral unbiasedness. 

We have extended this test to include varying widths in a two-layer neural network and obtained similar results.

\section{Robustness to an additive adversarial attack}
\label{Sec:Robustness}
This experiment compares the robustness of two-layer neural networks trained with ARFF and SGD against an additive adversarial black box attack. 

\textbf{Datasets}\\
We consider the MNIST dataset that consists of $70,000$ handwritten digit images of size 
$28 \times 28$ and its corresponding labels $y \in \{0, 1, ..., 9\}$. 
 The task is to classify the images between the digits $0$ to $9$.

We partition the datasets into training, validation, and test data 
with the corresponding ratio of $7:2:1$.

\textbf{Neural network structure and training}\\
We train ten two-layer neural networks $\boldsymbol{\beta}^i(\boldsymbol{x}; \boldsymbol{\omega}^i, 
    \boldsymbol{\hat{\beta}}^i, b^i) = \sum^K_{k=1} \hat{\beta}^i_k \cos
    ({\boldsymbol{\omega}_k^i}\cdot \boldsymbol{x} + b^i), \ i=0, ..., 9
    \label{eq: NN for classification}$ with $K=2^{10}$ nodes each, with both ARFF and SGD where $ \boldsymbol{\omega}^i_k \in \mathbb{R}^d, \boldsymbol{x} \in \mathbb{R}^d, b^i \in \mathbb{R}, \hat{\beta}^i_k\in\mathbb{R}$, 
and $d=784$ for the MNIST dataset. We train each neural network to classify one number each. E.g. $\boldsymbol{\beta}^2(\boldsymbol{x}; \boldsymbol{\omega}^2, 
    \boldsymbol{\hat{\beta}}^2, b^2)$ is trained to tell if a handwritten digit is the number $2$ or not. For SGD we use one least squares loss function for each neural network.

The training data $\{(\boldsymbol{x}_n;(y_n^0, y_n^1, 
..., y_n^9))\}^N_{n=1}$ 
consist of vectorized handwritten digit images $\boldsymbol{x}_n \in \mathbb{R}^{784}$ with corresponding vector labels $(y_n^0, y_n^1, 
..., y_n^9)$. Each vector label has one component equal to one and the others equal to zero. 
The index $i$ of the component $y_n^i$ equal to $1$ is the number that the handwritten digit $\boldsymbol{x}_n$ represents. 
    


In the following experiments, we run both the training algorithms, SGD and ARFF, for $M=100$ epochs each.
For ARFF, the parameters are set to  
$\lambda=0.05$, $\delta=0.02$  $\gamma=3d-2$, with $d=784$. For SGD, we use a batch size of $32$ and a learning rate of $0.002$. 

\textbf{Classification and computation of the accuracy}

A handwritten test digit $\boldsymbol{x}\in\mathbb{R}^{784}$ is classified as the number 
$\arg\max_i \{\boldsymbol{\beta}^i(\boldsymbol{x}; \boldsymbol{\omega}^i, 
    \boldsymbol{\hat{\beta}}^i,b^i)\}
    \label{eq: number classification criteria}$
where $\{\omega^i_k, \hat{\beta}_k^i, b_k^i\}_{k=1}^K, i=0,...,9$ 
are the trained weights and biases resulting from either ARFF or SGD. 

We evaluate the performance of the artificial neural networks 
by the classification accuracy, defined as $accuracy = \frac{N_{\text{correctly classified}}}{N_{\text{total}}}$, where $N_{\text{total}}$ represents the total number of test images, and $N_{\text{correctly classified}}$ denotes the count of accurately classified handwritten test images out of $N_{\text{total}}$.
\vspace{-3mm}
\subsection*{Random noise attack}
\vspace{-1mm}We apply the simple black box attack of adding random normal distributed noise to $n_{pixel}\in[0, d]$ randomly chosen pixels of each $N_{test}$ images in the test dataset to get the noisy test data $\boldsymbol{x}^{n}_{noise} = \boldsymbol{x}^{n}_{test} + \boldsymbol{\delta}^n \odot
    \boldsymbol{r}^n, $
$\boldsymbol{r}^n \sim N(\boldsymbol{0}, \sigma^2 \boldsymbol{I}), \;n = 1,..., N_{test}$ where $I\in\mathbb{R}^{d\times d}$ is the identity matrix, $\sigma\in\mathbb{R}$, and $\odot$ denotes component-wise multiplication. Here the component $\delta_i$ of $\boldsymbol{\delta}$ satisfies $\delta_i \in \{0, 1\}$ and the number of the elements which are equal to one is $n_{pixel}$, i.e. 
$\sum^d_{i=1} \delta_i = n_{pixel}$.

\begin{figure}[h]
	\begin{center}
  \centerline{\includegraphics[width = 0.34 \linewidth]{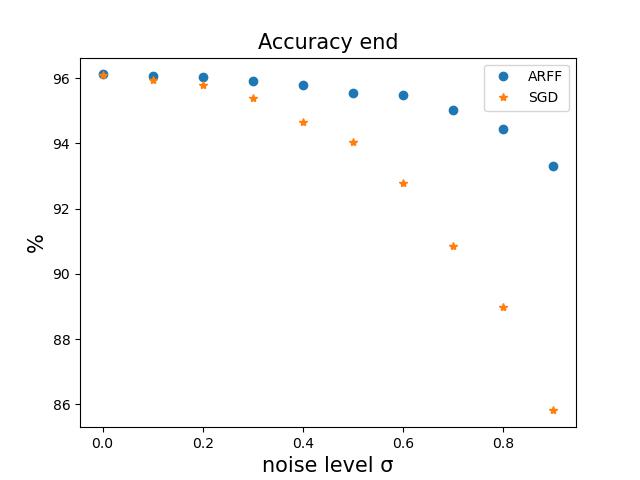}\includegraphics[width = 0.34 \linewidth]{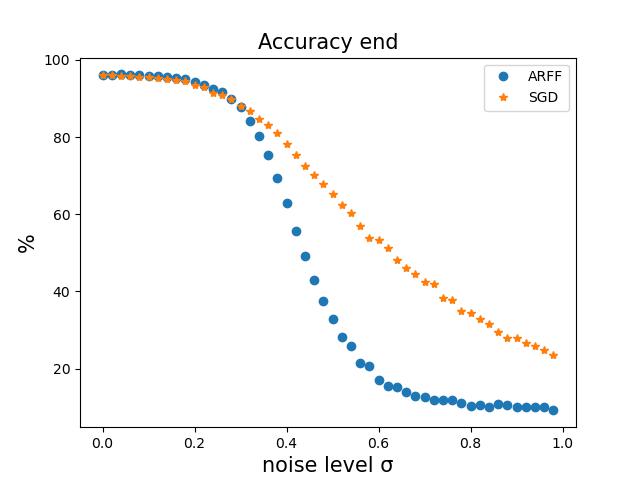}\includegraphics[width = 0.34 \linewidth]{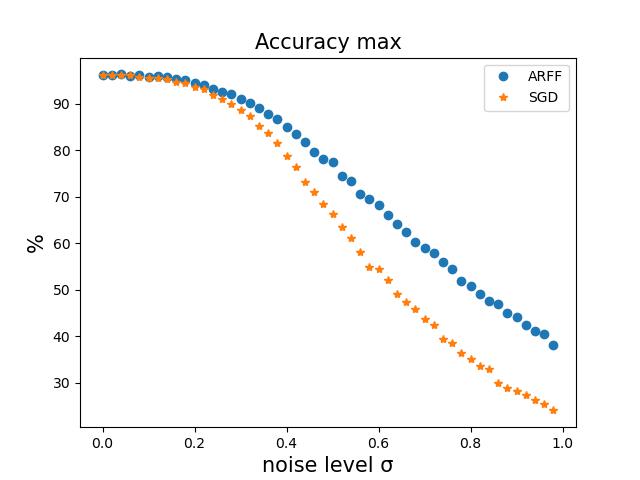}}
	\end{center}
	\caption{\textbf{Left, Experiment 1:} The figure shows the accuracy on the MNIST test data after a sparse black box attack.
 \textbf{Middle, Experiment 2:} Accuracy on the MNIST test data after a black box attack for different noise levels.
 \textbf{Right, Experiment 3:} Accuracy on the MNIST test dataset after a black box noise attack for classifiers tuned, with the help of noisy validation data, to withstand noise attacks. In Experiment 3 the stopping criteria depends on the noise level $\sigma$.}
	\label{fig:full_attack_both}
\end{figure}

\textbf{Experiment 1:}
 We present the classification rates on the noisy test images, where $n_{pixel} = 50$, for different values of $\sigma$ on the left in Figure \ref{fig:full_attack_both} for both SGD and ARFF. Even though both training algorithms have trained the neural networks to the same classification rate, the accuracy of the neural networks trained by SGD decreases faster when $\sigma$ increases compared to the accuracy of the neural networks trained by ARFF.
 
\textbf{Experiment 2:}
We present the classification rates on the noisy test images, where $n_{pixel} = d = 784$, for different values of $\sigma$ in the middle in Figure \ref{fig:full_attack_both}. Contrary to the results in Experiment 1, we see a faster decrease in the accuracy when $\sigma$ increases for ARFF than for SGD.

\textbf{Experiment 3:}
As in Experiment 2 we use $n_{pixel} = d = 784$. The difference is that we also add noise, from $\mathcal{N}(0, \sigma^2)$, to all $784$ pixels of each image in the validation dataset and use the best classification rate on the validation data as stopping criterion for the training. We present the results on the right in Figure \ref{fig:full_attack_both}. The classification rate for SGD is qualitatively the same as in Experiment 2, but now the classification rate for ARFF is better than for SGD. The optimal stopping epoch differs for each value of $\sigma$, though.

\textbf{Remark:}
For comparison, we run Experiments 1, 2, and 3 analogously on the CIFAR-10 dataset as well. The results are qualitatively the same. I.e., ARFF shows better results for Experiments 1 and 3, while SGD gives better results for Experiment 2.
\vspace*{-4.0 mm}
\section{Conclusion}
Our experimental findings underscore the effectiveness of ARFF-trained neural networks in efficiently capturing high-frequency content within the context of function reconstruction. We have also demonstrated the heightened robustness of ARFF-trained models when subjected to sparse black-box noise attacks at varying levels. While the same superiority is not consistently observed in the face of full black-box noise attacks, where noise is added to every pixel of the input image, we establish that the ARFF-trained neural network can maintain its robustness and outperform SGD by strategically leveraging noisy validation datasets and implementing early stopping. This insight provides practitioners with a viable strategy for prioritizing neural network robustness. Moreover, this study paves the way for future investigations, urging a more comprehensive exploration of ARFF training under sophisticated and well-structured attacks, the Square Attack \cite{andriushchenko2020square}, which we eagerly anticipate exploring in subsequent research endeavors.
\newpage

\section{Acknowledgement}
We express our gratitude to Prof. Anders Szepessy and Dr. Erik von Schwerin for reading the manuscript and providing their valuable feedback. This collaborative work, integral to Ms. Lisi Liang's master's thesis, is submitted as a component of her master's degree fulfillment at RWTH Aachen University.

We also acknowledge the financial support by the KAUST Office of Sponsored Research (OSR) under Award numbers URF$/1/2281-01-01$, URF$/1/2584-01-01$ in the KAUST Competitive Research Grants Program Round 8 and the Alexander von Humboldt Foundation.

\bibliographystyle{unsrt}  
\bibliography{references} 

\begin{thebibliography}{10}

\bibitem{Barron_1993}
Andrew~R. Barron.
\newblock Universal approximation bounds for superpositions of a sigmoidal
  function.
\newblock {\em IEEE Transactions on Information Theory}, 39(3):930--945, 1993.

\bibitem{kammonen2020adaptive}
Aku Kammonen, Jonas Kiessling, Petr Plech{\'a}{\v{c}}, Mattias Sandberg, and
  Anders Szepessy.
\newblock Adaptive random fourier features with metropolis sampling.
\newblock {\em Foundations of Data Science}, 2(3):309--332, 2020.

\bibitem{kiessling2021}
Jonas Kiessling, Emanuel Str{\"o}m, and Ra{\'u}l Tempone.
\newblock Wind field reconstruction with adaptive random fourier features.
\newblock {\em Proc. R. Soc. A.}, 477, 2021.

\bibitem{kammonen2022deepVSshallow}
Aku Kammonen, Jonas Kiessling, Petr Plech{\'a}{\v{c}}, Mattias Sandberg, Anders
  Szepessy, and Ra{\'u}l Tempone.
\newblock {Smaller generalization error derived for a deep residual neural
  network compared with shallow networks}.
\newblock {\em IMA Journal of Numerical Analysis}, 43(5):2585--2632, 09 2022.

\bibitem{rahimi2007random}
Ali Rahimi and Benjamin Recht.
\newblock Random features for large-scale kernel machines.
\newblock {\em Advances in neural information processing systems}, 20, 2007.

\bibitem{Vladimir1998}
Vapnik Vladimir~Naumovich.
\newblock {\em Statistical learning theory}.
\newblock Adaptive and learning systems for signal processing, communications,
  and control. Wiley, New York, 1998.

\bibitem{SCSpringer2008}
Ingo Steinwart and Andreas Christmann.
\newblock {\em Support Vector Machines}.
\newblock Springer Publishing Company, Incorporated, 1st edition, 2008.

\bibitem{tancik2020fourfeat}
Matthew Tancik, Pratul~P. Srinivasan, Ben Mildenhall, Sara Fridovich-Keil,
  Nithin Raghavan, Utkarsh Singhal, Ravi Ramamoorthi, Jonathan~T. Barron, and
  Ren Ng.
\newblock Fourier features let networks learn high frequency functions in low
  dimensional domains.
\newblock {\em NeurIPS}, 2020.

\bibitem{kiessling2022computable}
Jonas Kiessling and Filip Thor.
\newblock A computable definition of the spectral bias.
\newblock In {\em Proceedings of the AAAI Conference on Artificial
  Intelligence}, volume~36, pages 7168--7175, 2022.

\bibitem{hajri2022neural}
Hatem Hajri, Manon Cesaire, Lucas Schott, Sylvain Lamprier, and Patrick
  Gallinari.
\newblock Neural adversarial attacks with random noises.
\newblock {\em International Journal on Artificial Intelligence Tools}, 2022.

\bibitem{tsai2019adversarial}
Yi-Ting Tsai, Min-Chu Yang, and Han-Yu Chen.
\newblock Adversarial attack on sentiment classification.
\newblock In {\em Proceedings of the 2019 ACL Workshop BlackboxNLP: Analyzing
  and Interpreting Neural Networks for NLP}, pages 233--240, 2019.

\bibitem{andriushchenko2020square}
Maksym Andriushchenko, Francesco Croce, Nicolas Flammarion, and Matthias Hein.
\newblock Square attack: a query-efficient black-box adversarial attack via
  random search.
\newblock In {\em European conference on computer vision}, pages 484--501.
  Springer, 2020.

\bibitem{Friendly_noise_against_adversarial_noise}
Tian~Yu Liu, Yu~Yang, and Baharan Mirzasoleiman.
\newblock Friendly noise against adversarial noise: A powerful defense against
  data poisoning attack.
\newblock In S.~Koyejo, S.~Mohamed, A.~Agarwal, D.~Belgrave, K.~Cho, and A.~Oh,
  editors, {\em Advances in Neural Information Processing Systems}, volume~35,
  pages 11947--11959. Curran Associates, Inc., 2022.

\bibitem{Abadi2016DeepLW}
Mart{\'i}n Abadi, Andy Chu, Ian~J. Goodfellow, H.~B. McMahan, Ilya Mironov,
  Kunal Talwar, and Li~Zhang.
\newblock Deep learning with differential privacy.
\newblock {\em Proceedings of the 2016 ACM SIGSAC Conference on Computer and
  Communications Security}, 2016.

\bibitem{Chen2018DetectingBA}
Bryant Chen, Wilka Carvalho, Nathalie Baracaldo, Heiko Ludwig, Ben Edwards,
  Taesung Lee, Ian Molloy, and B.~Srivastava.
\newblock Detecting backdoor attacks on deep neural networks by activation
  clustering.
\newblock {\em ArXiv}, abs/1811.03728, 2018.

\bibitem{rahaman2019spectral}
Nasim Rahaman, Aristide Baratin, Devansh Arpit, Felix Draxler, Min Lin, Fred
  Hamprecht, Yoshua Bengio, and Aaron Courville.
\newblock On the spectral bias of neural networks.
\newblock In {\em International Conference on Machine Learning}, pages
  5301--5310. PMLR, 2019.

\bibitem{Hong2022OnTA}
Qingguo Hong, Qinyan Tan, Jonathan~W. Siegel, and Jinchao Xu.
\newblock On the activation function dependence of the spectral bias of neural
  networks.
\newblock {\em ArXiv}, abs/2208.04924, 2022.

\bibitem{Wu2022ExtrapolationAS}
Yongtao Wu, Zhenyu Zhu, Fanghui Liu, Grigorios~G. Chrysos, and Volkan Cevher.
\newblock Extrapolation and spectral bias of neural nets with hadamard product:
  a polynomial net study.
\newblock {\em ArXiv}, abs/2209.07736, 2022.

\bibitem{Rahaman2018OnTS}
Nasim Rahaman, Devansh Arpit, Aristide Baratin, Felix Dr{\"a}xler, Min Lin,
  Fred~A. Hamprecht, Yoshua Bengio, and Aaron~C. Courville.
\newblock On the spectral bias of deep neural networks.
\newblock {\em ArXiv}, abs/1806.08734, 2018.

\bibitem{cao2019towards}
Yuan Cao, Zhiying Fang, Yue Wu, Ding-Xuan Zhou, and Quanquan Gu.
\newblock Towards understanding the spectral bias of deep learning.
\newblock {\em arXiv preprint arXiv:1912.01198}, 2019.

\bibitem{xu2019frequency}
Zhi-Qin~John Xu, Yaoyu Zhang, Tao Luo, Yanyang Xiao, and Zheng Ma.
\newblock Frequency principle: Fourier analysis sheds light on deep neural
  networks.
\newblock {\em arXiv preprint arXiv:1901.06523}, 2019.

\bibitem{xu2019training}
Zhi-Qin~John Xu, Yaoyu Zhang, and Yanyang Xiao.
\newblock Training behavior of deep neural network in frequency domain.
\newblock In {\em Neural Information Processing: 26th International Conference,
  ICONIP 2019, Sydney, NSW, Australia, December 12--15, 2019, Proceedings, Part
  I 26}, pages 264--274. Springer, 2019.

\bibitem{weinan2019comparative}
E~Weinan, Chao Ma, and Lei Wu.
\newblock A comparative analysis of optimization and generalization properties
  of two-layer neural network and random feature models under gradient descent
  dynamics.
\newblock {\em Sci. China Math}, 2020.

\bibitem{BarronSpace-Weinan2022}
E~Weinan, Chao Ma, and Lei Wu.
\newblock The barron space and the flow-induced function spaces for neural
  network models.
\newblock {\em Constructive Approximation}, 55(1):369--406, February 2022.

\bibitem{ruder2016overview}
Sebastian Ruder.
\newblock An overview of gradient descent optimization algorithms.
\newblock {\em arXiv preprint arXiv:1609.04747}, 2016.

\bibitem{duchi2011adaptive}
John Duchi, Elad Hazan, and Yoram Singer.
\newblock Adaptive subgradient methods for online learning and stochastic
  optimization.
\newblock {\em Journal of machine learning research}, 12(7), 2011.

\bibitem{kingma2014adam}
Diederik~P Kingma and Jimmy Ba.
\newblock Adam: A method for stochastic optimization.
\newblock {\em arXiv preprint arXiv:1412.6980}, 2014.

\bibitem{rudi2017generalization}
Alessandro Rudi and Lorenzo Rosasco.
\newblock Generalization properties of learning with random features.
\newblock {\em Advances in neural information processing systems}, 30, 2017.

\bibitem{gurney1997introduction}
Kevin Gurney.
\newblock {\em An introduction to neural networks}.
\newblock CRC press, 1997.

\bibitem{shalev2014understanding}
Shai Shalev-Shwartz and Shai Ben-David.
\newblock {\em Understanding machine learning: From theory to algorithms}.
\newblock Cambridge university press, 2014.

\bibitem{bishop1994neural}
Chris~M Bishop.
\newblock Neural networks and their applications.
\newblock {\em Review of scientific instruments}, 65(6):1803--1832, 1994.

\bibitem{szegedy2013intriguing}
Christian Szegedy, Wojciech Zaremba, Ilya Sutskever, Joan Bruna, Dumitru Erhan,
  Ian Goodfellow, and Rob Fergus.
\newblock Intriguing properties of neural networks.
\newblock {\em arXiv preprint arXiv:1312.6199}, 2013.

\bibitem{carlini2017towards}
Nicholas Carlini and David Wagner.
\newblock Towards evaluating the robustness of neural networks.
\newblock In {\em 2017 ieee symposium on security and privacy (sp)}, pages
  39--57. Ieee, 2017.

\bibitem{goodfellow2014explaining}
Ian~J Goodfellow, Jonathon Shlens, and Christian Szegedy.
\newblock Explaining and harnessing adversarial examples.
\newblock {\em arXiv preprint arXiv:1412.6572}, 2014.

\end{thebibliography}
\nocite{weinan2019comparative}
\nocite{BarronSpace-Weinan2022}
\nocite{ruder2016overview, duchi2011adaptive, kingma2014adam, rudi2017generalization}
\nocite{gurney1997introduction, shalev2014understanding, bishop1994neural}
\nocite{szegedy2013intriguing, carlini2017towards, goodfellow2014explaining}
\nocite{hajri2022neural}
\end{document}